\documentclass[final,5p,twocolumn]{elsarticle}

\usepackage{amssymb}
\usepackage{amsthm}

\journal{Springer}

\begin{document}

\begin{frontmatter}



\title{Web page classification with Google Image Search results}

\author[1]{Fahri Aydos}
\author[1]{A. Murat \"{O}zbayo\u{g}lu}
\author[2]{Yahya \c{S}irin}
\author[1]{M. Fatih Demirci}

\address[1]{TOBB University of Economics \& Technologies, Ankara, Turkey}
\address[2]{\.{I}stanbul Sabahattin Zaim University, \.{I}stanbul, Turkey}

\begin{abstract}
    In this paper, we introduce a novel method that combines multiple neural network results to decide the class of the input. This is the first study which used the method for web pages classification. In our model, each element is represented by multiple descriptive images. After the training process of the neural network model, each element is classified by calculating its descriptive image results. We apply our idea to the web page classification problem using Google Image Search results as descriptive images. We obtained a classification rate of 94.90\% on the WebScreenshots dataset that contains 20000 web sites in 4 classes. The method is easily applicable to similar problems.
\end{abstract}




\end{frontmatter}


\section{Introduction}\label{sec:introduction}

Web page classification has been a substantial problem since the early days of the Internet \cite{hashemi2020web}. In addition to the exponential growth of the Internet in size, new technologies and usage areas are developed day by day. With these developments in the Internet, users, companies, and governments need productive and efficient ways to analyze and classify web pages and other Internet services. According to the needs of the classifier, web pages are classified mainly by their content topic, whether they are adult, fraud, or offensive websites. There are three significant difficulties in web page classification. The first is the continuous addition of new content to the Internet. Second, web pages have many attributes that can be used for classification. This makes it difficult to get a standard when classifying and requires the usage of complex techniques. The third is to find adequate and descriptive datasets.

Web pages have many attributes, such as URL address, text content, hyperlinks, image content, domain and server information, HTML tags, semantic web tags. There are many studies to classify web sites with one or many of these features. A web page is basically a text file where content and design are combined using HTML codes. Therefore text classification methods are dominant in the literature. But separating content from design, detecting the language of the page, and cleaning unnecessary words requires attention and complex analysis.

With this paper, we introduce a simple method to classify web pages with images. Since deep learning is successful with image based classifications, we built our method on deep learning algorithms.

Deep learning methods have become more popular and applied to supervised, semi-supervised, unsupervised, and reinforcement learning problems in the last decade. Especially convolutional neural networks (CNN) show the enormous success in visual problems, such as object detection, classification, face recognition. In 1989 LeCun et al. used a neural network with 3 hidden layers to recognize handwritten digits \cite{lecun1989backpropagation}. Over time, computers become more powerful, and GPU usage improved. As a result, new architectures of CNN developed for different problems with having different input analysis styles. Transfer learning was also started to be used to save time and operation costs.

In our method, every web site is described by more than one image, called ``descriptive images" for the web site. Our descriptive images are the Google Image Search results for web pages URL addresses. We trained well known CNN  architectures with these descriptive images. Our test web sites have descriptive images also. We get CNN results for each image and combine these results into one result with specific metrics and define the test web site's class.

To find an adequate dataset is an essential problem for all Machine Learning researches. This is also true for Internet related problems. Internet content can change very quickly, and a web site can be closed, parked, or change topic. Most of the web site datasets are small and cannot represent the Internet to the desired extent. Some of them can stay old and require renewal.

With this paper, we also introduce our WebScreenshots dataset, that is suitable for content based or screenshot based web site classifications. WebScreenshots contains 20000 web sites in 4 classes.

The outline of this paper is the following: In Section \ref{sec:relatedwork}, we provide a literature review for web page classification with visual features. In Section \ref{sec:dataset}, we introduce our WebScreenshots dataset. In Section \ref{sec:method}, we declare our method and metrics used in the method. In Section \ref{results}, we explain our experiments, state the results, compare and discuss these results with other methods in the literature. Finally, in Section \ref{results}, we end up with a summary.


\section{Related work}\label{sec:relatedwork}

Web pages have different kinds of features. Hashemi divided the studies in the web pages classification field into three groups: Text-based, image-based, and combined usage of text-based and image-based features \cite{hashemi2020web}. Despite Hashemi, we divided into four main groups according to what features are used as classifier: (1) Textual classifications: URL address, text content, title, HTML description, HTML code, etc. (2) Visual classifications: images, design, videos, etc. (3) Graph-based classifications: hyperlink structures, neighbor web sites. (4) And other information: user behaviors, web directories, semantic web, raw data of domain (IP address, owner, hosting server, hosting country). As Hashemi, we can add combined methods as another group.

Our papers lies in image-based, visual classfification group. In image based classifications, there are four image sources for a web page: (1) screenshots of the web page, (2) image and video content on the web page, (3) preclassified images, and (4) related images to the web page. Below we will examine the use of these sources in the literature.

\subsection{Web page screenshots}\label{sec:screenshots}


In screenshot based classifications, the screenshot of the web page is used as the descriptor of the web page. Mirdehghani and Monadjemi analyzed web page screenshots to automatically evaluate the web page's aesthetic quality \cite{mirdehghani2009web}. They extracted color space histogram and Gabor features and fed them to Artificial Neural Networks (ANN). They used university web sites as the dataset. De Boer et al. used the same method and added Edge histogram and Tamura features \cite{deboer2010classifying}. Along with aesthetic quality, they also tried to classify recency and the topic of web pages. Their dataset consisted of 120 web pages. Videira and Goncalves improved the method by feature selection techniques and applied Bag of Visual Words to SIFT descriptors \cite{videira2014automatic}. 82,50\% accuracy was achieved for four classes, and 63.75\% for eight classes. Their dataset consisted of 90 web pages for each class. In 2019 Dou et al. rated web pages for their aesthetics value with deep learning methods \cite{dou2019webthetics}, using a dataset that contained 398 web page screenshots. They build a neural network architecture and used transfer learning. They made a comparison with user's ratings for the web pages with their deep learning model results.

\subsection{Image content}\label{sec:imagecontent}

Another method to classify web sites is categorizing a web page with the class of its image content. In this method, the images on the web page are analyzed with image processing techniques, and the web page is classified with the class of the statically dominant images. This method is mostly used for binary classification problems such as detecting pornographic web pages. Arentz and Olstad extracted image features from skin regions and used a genetic algorithm to train their system \cite{arentz2004classifying}. Rowley et al. used the same method and added clutter features and face detection techniques \cite{rowley2006large}. Hammami et al. improved previous works by combining textual and visual analyzing results \cite{hammami2005webguard}. Their method was to use the ratio of skin-color pixels to all pixels. Hu et al. made a similar approach and added contour-based features to detect skin areas \cite{hu2007recognition}, Ahmadi et al. added shape based characteristics \cite{ahmadi2011intelligent} to previous works. Dong et al. combined the bag of visual words features of the images and textual information of the image and the web site to detect pornographic web sites \cite{dong2014adult}.

\subsection{Preclassified images}\label{sec:preclass}

Hashemi and Hall used image based classification to determine dark propaganda of violent extremist organizations with the help of deep learning methods \cite{hashemi2019detecting}. To realize their project, they built a dataset that contained 120000, manually classified images. They spent four years for building their dataset. To build the dataset, they used images from social media besides web pages. The downside of the dataset was that it was only related to one extreme group. For another threat, the dataset must be renewed with related images. They used AlexNet \cite{krizhevsky2012imagenet} architecture for CNN, and their overall generalization accuracy for eight classes was 86.08\%.


\subsection{Related images}\label{sec:preclass}

As far as we know, this is the first study in the literature that uses related pictures to classify web pages. We use Google Image Search results as the related images for the web pages. So we don't need to build an image dataset as Hashemi and Hall did \cite{hashemi2019detecting}. In their dataset, images are directly related to the class. In our case, we use images related to the item in the class, and we don't need to pay too much attention to how they describe the item.

\section{Dataset}\label{sec:dataset}

\subsection{WebScreenshots dataset}

As can be seen in Section \ref{sec:relatedwork}, researchers using images in web page classification have created their datasets. Since existing databases are small and do not contain descriptive images for the web pages, we also needed to create a new dataset, and build our WebScreenshots \cite{aydos2020web} dataset.

The main reason for a new dataset was to get up-to-date images from Google Image Search when querying web pages in the dataset. These current images would be related to the web page and its class for a new dataset. For an old dataset, Google Image Search results would be significantly irrelevant as web pages could be modified or deleted.

As soon as the WebScreenshot dataset is created, we made queries with Google Image Search and saved the result images for each web page in the dataset. So we got the up-to-date related pictures for web pages. WebScreenshot is a generic dataset that contains URLs, classes, text content, and screenshots, and can also be used in web page classification studies that are unrelated to our method. We also used screenshots of the dataset for method comparison. But we should point out that Google Image Search results are not part of the WebScreenshots dataset.

\begin{table*}[t]
    \centering
    \caption{WebScreenshots dataset distribution by web site language}
    \label{tab:languages}
    \begin{tabular*}{\textwidth}{lrlrlrlr}
        \noalign{\smallskip}
        \hline
        \noalign{\smallskip}
        English	&	9373	&	Turkish	&	312	&	Estonian	&	34	&	Tagalog	&	8	\\
        German	&	3086	&	Catalan	&	184	&	Ukrainian	&	29	&	Somali	&	8	\\
        French	&	1165	&	Swedish	&	184	&	Welsh	&	27	&	Bengali	&	6	\\
        Russian	&	1125	&	Vietnamese	&	118	&	Croatian	&	24	&	Thai	&	6	\\
        Italian	&	1085	&	Czech	&	98	&	Chinese	&	21	&	Albanian	&	4	\\
        Dutch	&	679	&	Norwegian	&	79	&	Korean	&	18	&	Latvian	&	4	\\
        Danish	&	471	&	Romanian	&	56	&	Lithuanian	&	17	&	Persian	&	3	\\
        Polish	&	455	&	Finnish	&	55	&	Indonesian	&	16	&	Swahili	&	3	\\
        Portuguese	&	392	&	Hungarian	&	39	&	Slovak	&	14	&	Afrikaans	&	3	\\
        Japanese	&	385	&	Bulgarian	&	37	&	Arabic	&	10	&	Slovenian	&	2	\\
        Spanish	&	320	&	Greek	&	35	&	Hebrew	&	9	&	Macedonian	&	1	\\
        \noalign{\smallskip}
        \hline
    \end{tabular*}
\end{table*}

To build WebScreenshots dataset, we chose our classes as ``machinery", ``music", ``sport", ``tourism" for the first step. Then we filtered Parsed DMOZ \cite{dataDMOZ} sites for each class. The Parsed DMOZ dataset has a directory structure. For example, a music site may be found under Japanese or Turkish directory. After filtering all web pages for a class, we took their screenshots. Following the removal of unreachable, parked, or closed web sites from the list, for the last step, we manually checked if the screenshot visually matches its class through the human eye. The web sites which could not visually identify their classes are eliminated. Considering 5000 web sites to be a good representative, we chose 5000 web sites from the list, and we did the same processes with the next class. For building a dataset with 20000 images, more than 50000 screenshots were needed to be processed manually.

Every web site in the dataset has the features: URL address, class, screenshot image, and text content. WebScreenshots dataset does not contain Google Image Search results due to copyright restrictions.

WebScreenshots is a multilingual dataset. A Python language detecting package found 44 languages in the dataset. These languages and the number of web pages in that language are shown in Table \ref{tab:languages}.

In our work, the dataset is divided into train, validation, and test sets by randomly chosen 4000, 500, and 500 web pages, respectively. To facilitate some process, a subset of the dataset (called Subset in the paper) was also randomly chosen with 500 web pages in each class (400 train, 50 validation, 50 test) with a total of 2000 web pages.

\begin{table}[t]
    \centering
    \caption{Number of images in image sets for WebScreenshots dataset}
    \label{tab:images}
    \small
    \begin{tabular}{lccc}
        \noalign{\smallskip}
        \hline
        \noalign{\smallskip}
        Image set   & \# of Sites & Train+Validation & Test \\
        \noalign{\smallskip}
        \hline
        \noalign{\smallskip}
        Google10    &     20,000  &  175,988 & 37,960 \\
        Google20    &     20,000  &  342,358 & 37,960 \\
        Subset10    &      2,000  &   17,550 &  3,851 \\
        Subset20    &      2,000  &   34,038 &  3,851 \\
        \noalign{\smallskip}
        \hline
    \end{tabular}
\end{table}

\subsection{Descriptive images: Google Image Search results}

\begin{figure*}[t]
    \includegraphics[width=\textwidth]{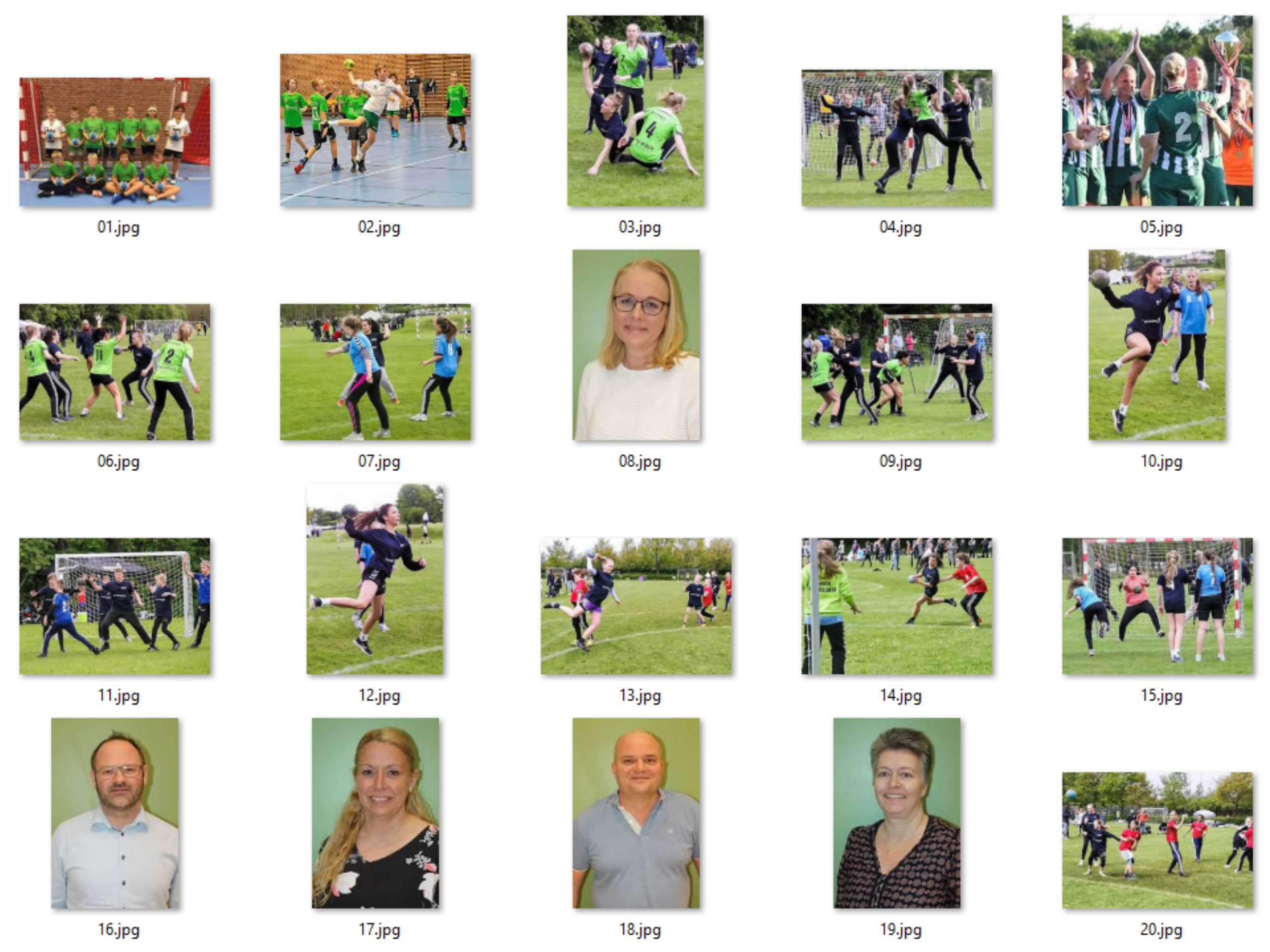}
    \caption{A sample site's results on Google Image Search}
    \label{fig:samplesite}
\end{figure*}

In our method, every element in the set is described by more than one image. We called these images as descriptive images. For the web site classification problem, there are some possibilities to define a web site with images: Screenshots or image content of the page, screenshots of neighbor web sites, search engine results, etc., could be used. To define a web site, we chose Google Image Search\footnote{https://www.google.com/imghp} results. We searched the URL address of the web page on Google Image Search and saved the first 20 images into a directory named after the web page. We filtered only photo-like images; vector graphics and social media icons are ignored. We also preserved the order of results and saved images consecutively with names as 01.jpg, 02.jpg, ..., 20.jpg. Some searches returned less than 20 images. We used the Google thumbnail images, stored on the Google Inc. servers since to get the original images from web sites needs more network bandwidth and time in addition to the difficulties such as could not be found, different format, etc.

A search result sample for a web site can be seen in Figure \ref{fig:samplesite}. For this web site, there are five portraits of employees. Even a human can not distinguish the class of the web site comparing only these images. A mechanical, music, sports, or tourism site can have images like these. But with the other fifteen images, one can easily classify this web site as a sports site.

We created four different descriptive image sets to visually describe web pages in the dataset: Google10, Google20, Subset10, and Subset20. Google10 and Google20 are for the full WebScreenshots dataset; Subset10 and Subset20 are for the subset of the dataset. Google10 and Subset10 were built with the first 10 images, and Google20 and Subset20 were built with all 20 images of the image search results.

The number of images in each image set can be seen in Table \ref{tab:images}. The reason for total numbers being less than expected values is that some searches returned less than 20 images. For test sets, we used all 20 images since we were able to control image numbers during the tests. When we have a trained architecture, we can use any number of images to test a web site, and we can feed the model with 5, 10, 15, or 20 images of the tested web site. More information can be found in Section \ref{sec:method} about this method.

\begin{figure*}[t]
    \includegraphics[width=\textwidth]{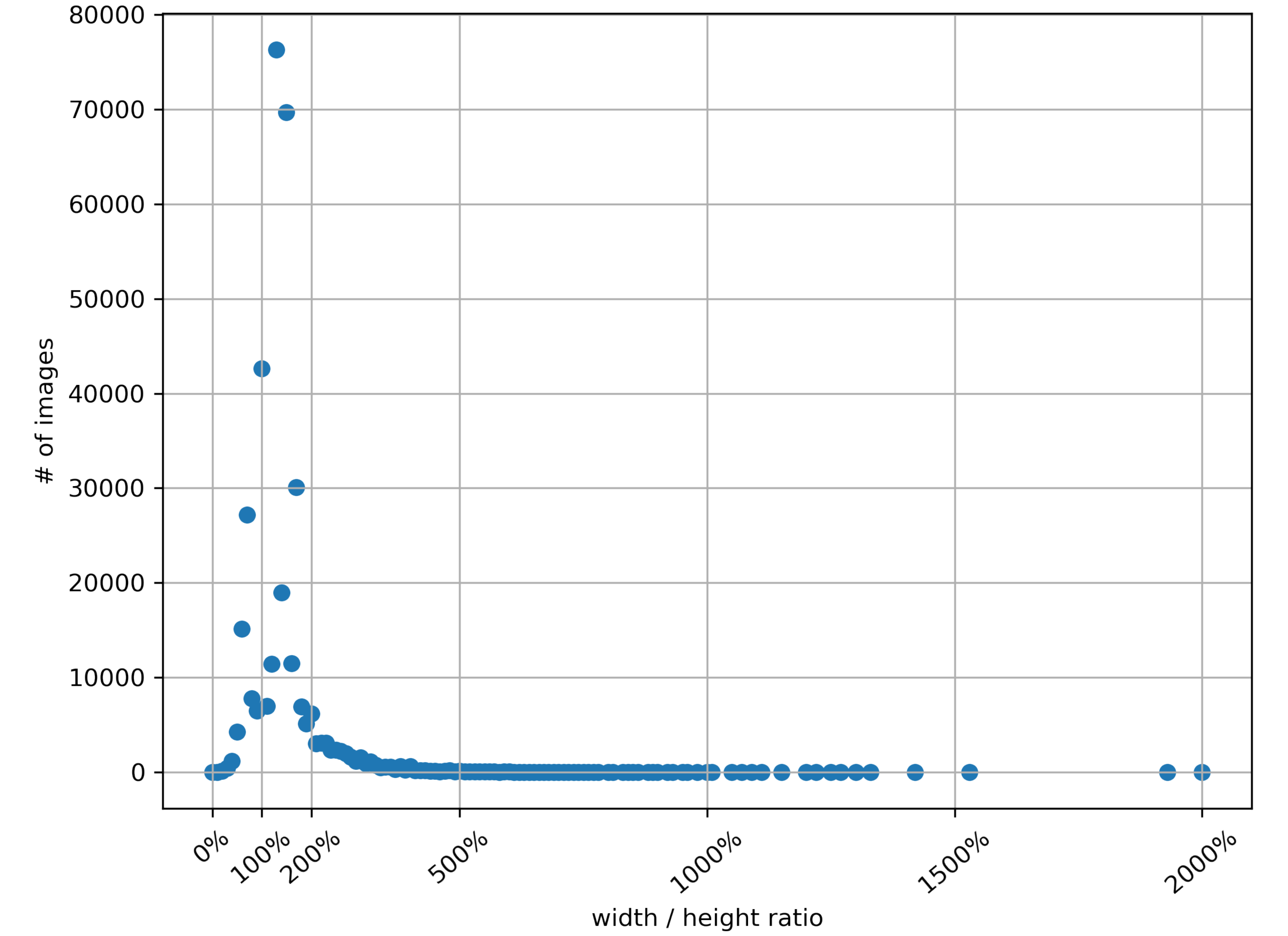}
    \caption{Histogram of width-height ratio of Google Image Search results.}
    \label{fig:whr}
\end{figure*}

The histogram of the width-height ratio of Google Image Search results are shown in Figure \ref{fig:whr}. The scale of the histogram is 10\%, and ratio labels show the base ratio (i.e., 0\% shows the value of 0\%-9\%). We discuss the effects of the information in this chart on our results on Section \ref{sec:results}.

\section{Proposed method}\label{sec:method}

Our method depends on obtaining all the results for descriptive images of a web site, combining them into one value, and getting the web site class. We used 3 kinds of accuracy metrics to get one result for a web page from its multiple descriptive images, and each metric has 4 levels. These metrics are based on the descriptive image's prediction values for each class. Since we have 4 classes, CNN architectures return 4 real numbers between 0 and 1. And since a web site has a maximum of 20 descriptive images, we have a result matrix with the dimensions of 4x20 for a web site. Our metrics are described below.

\textit{Summation (S) metrics} are calculated by summation of columns and the acceptance of the maximum valued column as the prediction of the web site. We made summations for the first 5, 10, 15, and all 20 rows. These calculations are abbreviated as S05, S10, S15, and S20 in the paper.

\textit{One-hot (H) metrics:} In the result matrix, we replaced the maximum valued class with 1, and the others with 0 for each descriptive image.  The same summation processes are repeated for the new one-hot matrix. One-hot calculations are abbreviated as H05, H10, H15, and H20 in the paper.

\textit{Average reordered (A) metrics:} We calculated the average of each column and reordered the result matrix via the column, which has the maximum average, from high value to lower. The same summation processes are repeated for the new reordered matrix. Average reordered calculations are abbreviated as A05, A10, A15, and A20 in the paper. Note that, since the calculation of S20 and A20 are the same, they have the same results.

A sample result matrix for one web site is presented in Table \ref{tab:sampleresults}. One-hot and averaged reordered transformations of this matrix can be seen on Table \ref{tab:onehot} and Table \ref{tab:averagereordered} respectively. Notice that after the average reordered transformation, the Google result order is changed. This shows us that Google Image Search results are related to the search query itself, rather than the class of web site on the search query.

Let $W=\{w_1,w_2,...,w_t\}$ be the all web sites in the test set, $I_{i,c}^{w}$ be the CNN results matrix of descriptive images of a web page $w$ with $i\in\{1,2,..,20\}$, $c\in\{1=machinery,2=music,3=sport,4=tourism\}$, $\bar{I}_{i,c}^{w}$ be the one-hot matrix, and $\hat{I}_{i,c}^{w}$ be the average reordered matrix of the $I_{i,c}^{w}$ matrix. Summation, one-hot and average reordered results for one web page $w$ and for image count $k$, can be formulate respectively as below:

\begin{equation}
    S_{k,c}^{w} = \sum_{i=1}^{k} I_{i,c}
\end{equation}
\begin{equation}
    H_{k,c}^{w} = \sum_{i=1}^{k} \bar{I}_{i,c}
\end{equation}
\begin{equation}
    A_{k,c}^{w} = \sum_{i=1}^{k} \hat{I}_{i,c}
\end{equation}

In our work, we study 5, 10, 15, and 20 as the value of $k$. Maximum valued class (e.g.: $CS_{k}^{w}=max(S_{k,1}^{w},S_{k,2}^{w},S_{k,3}^{w},S_{k,4}^{w})$) is defined as the predicted class of the web site. Accuracies are calculated as the ratio of correctly predicted web sites to the total number of web sites in the test set. For example, for the summation metric:

\begin{equation}
    S05 = \frac{\#\ of\ correct\ prediction\ with\ k=5}{\#\ of\ web\ sites\ in\ test\ set}
\end{equation}

We did the same calculations even if a web page has less than 20 descriptive images. So for example, if a web page has 8 images, S10, S15, and S20 have the same value. We also calculated the per image accuracy, the accuracy of every image individually in the test sets. There are 13 accuracy metrics in total.

\section{Experiments and results}\label{results}

\subsection{Experiments}

In our experiments, we used 6 different CNN architectures: VGG16 \cite{simonyan2014very}, DenseNet121, DenseNet169, DenseNet201 \cite{huang2017densely}, RestNet50, and InceptionRestNetV2 \cite{szegedy2017inception}. These are well known CNN architectures and used in many problems. For example, VGG16 won the ImageNet Large Scale Visual Recognition Challenge (ILSVRC) in 2014 and ResNet won in 2015 \cite{russakovsky2015imagenet}. We also took advantage of transfer learning from the ImageNet dataset. ImageNet is a dataset containing 1000 classes and over 1.2 million natural images \cite{deng2009imagenet}.

Having built the CNN architecture, we replaced the weights with pre-trained ImageNet weights. Each architecture had different pre-trained ImageNet weights. Since we have 4 classes instead of 1000 classes, we replaced the last dense layer with a layer for 4 classes. We trained only the last two layers: last dense layer and the previous average pooling layer. We used categorical cross-entropy as the loss function, and we set up the learning rate to 0.00001. Since popular deep learning frameworks contain all used CNN architectures and their ImageNet weights for transfer learning, any researcher can easily build up our experimental setup.

During the neural network training process, validation accuracy gives only per image success. Our metrics must be applied to the results of the test sets. So we saved the network architectures on every five epochs and run our metrics on these saved architectures with test sets. Figure \ref{fig:metrics} shows the all metrics success for best model Google10 \& DenseNet169. We trained the neural networks up to 100 epochs.

\subsection{Results}\label{sec:results}

\begin{table*}[t]
    \centering
    \caption{Method accuracies and comparison with other methods}
    \label{tab:success}
    \resizebox{\textwidth}{!}{
        \begin{tabular}{lcccccc}
            \noalign{\smallskip}
            \hline
            \noalign{\smallskip}
            Train set    & Text C. & BoVW    & VGG16   & DenseNet121 & DenseNet169 & DenseNet201 \\
            \noalign{\smallskip}
            \hline
            \noalign{\smallskip}
            Page Contents& \%92.65 &    -    &    -    &    -    &    -    &    -    \\
            Screenshots  &    -    & 46.95\% & 59.60\% & 60.55\% & 68.10\% & 68.40\% \\
            \noalign{\smallskip}
            \hline
            \noalign{\smallskip}
            Subset10     &    -    & 62.50\% & 81.00\% & 95.00\% & 97.00\% & 96.50\% \\
            Subset20     &    -    & 61.50\% & 83.50\% & 94.00\% & 96.50\% & 97.50\% \\
            Google10     &    -    &    -    & 89.55\% & 93.30\% & 94.90\% & 94.35\% \\
            Google20     &    -    &    -    & 90.35\% & 93.00\% & 94.75\% & 94.45\% \\
            \noalign{\smallskip}
            \hline
        \end{tabular}
    }
\end{table*}

The chosen CNN architectures gave different accuracies since they had different characteristics in analyzing the input images. With VGG16, we get reasonable results, but DenseNets are better than VGG16. DenseNet169 and DenseNet201 have maximum success rates. Validation accuracy and loss of RestNet50 and InceptionRestNetV2 were not stable even with a small learning rate, so we excluded them from our results. We used these architectures with their trained models with ImageNet dataset. We can say that trained model of VGG16 is not good enough for our dataset. Both ResNets and DenseNets are successful on extracting features from ImageNet dataset. But we used Google Image Search thumbnails images on our models and there were two major differences between our dataset and ImageNet. Firstly our dataset consisted of relatively small images, most of them have less width or height than the expected input of the CNN architecture’s default value, which is 224 pixels. Secondly ImageNet focused on objects, and in our dataset there were pictures of people, landscapes, events and designs. Densenets had been shown to be successful in this regard.

The chosen CNN architectures gave different accuracies since they had different characteristics in analyzing the input images. With VGG16, we get reasonable results, but DenseNets are better than VGG16. DenseNet169 and DenseNet201 have maximum success rates. RestNet50 and InceptionRestNetV2 have high success rates, but validation accuracy and loss are not stable even with a small learning rate. This may be observed because of the inputs that we had used. Google Image Search thumbnail images are small, and most of them have less width or height than the expected input of the CNN model’s default value, which is 224 pixels. RestNet50 and InceptionRestNetV2 may work for other problems and/or inputs. We excluded them from our results.

This may be observed because of the inputs that we had used. Google Image Search thumbnail images are small, and most of them have less width or height than the expected input of the CNN model’s default value, which is 224 pixels. RestNet50 and InceptionRestNetV2 may work for other problems and/or inputs. We excluded them from our results.

Our method's accuracies and comparison with other methods in the literature are shown in Table \ref{tab:success}. Our method values are on the last four rows and the last four columns. Page Content, Screenshots, and Bag of Visual Words (BoVW) values are accuracies of other methods. All experiments run on the WebScreenshots dataset.

Experiments with Subset10 and Subset20, which contain 2000 web sites, have a success rate of 97.50\%. But we prefer the success rate of the full dataset (Google10 and Google20) since a dataset with 20000 web sites is a better representation for the Internet. The maximum success rate of 94.90\% is obtained with DenseNet169 and Google10 descriptive images. But we observed close values with all Google10, Google20, and DenseNet169, DenseNet201 models.

\begin{figure*}[t]
    \centering
    \includegraphics[width=1.0\textwidth]{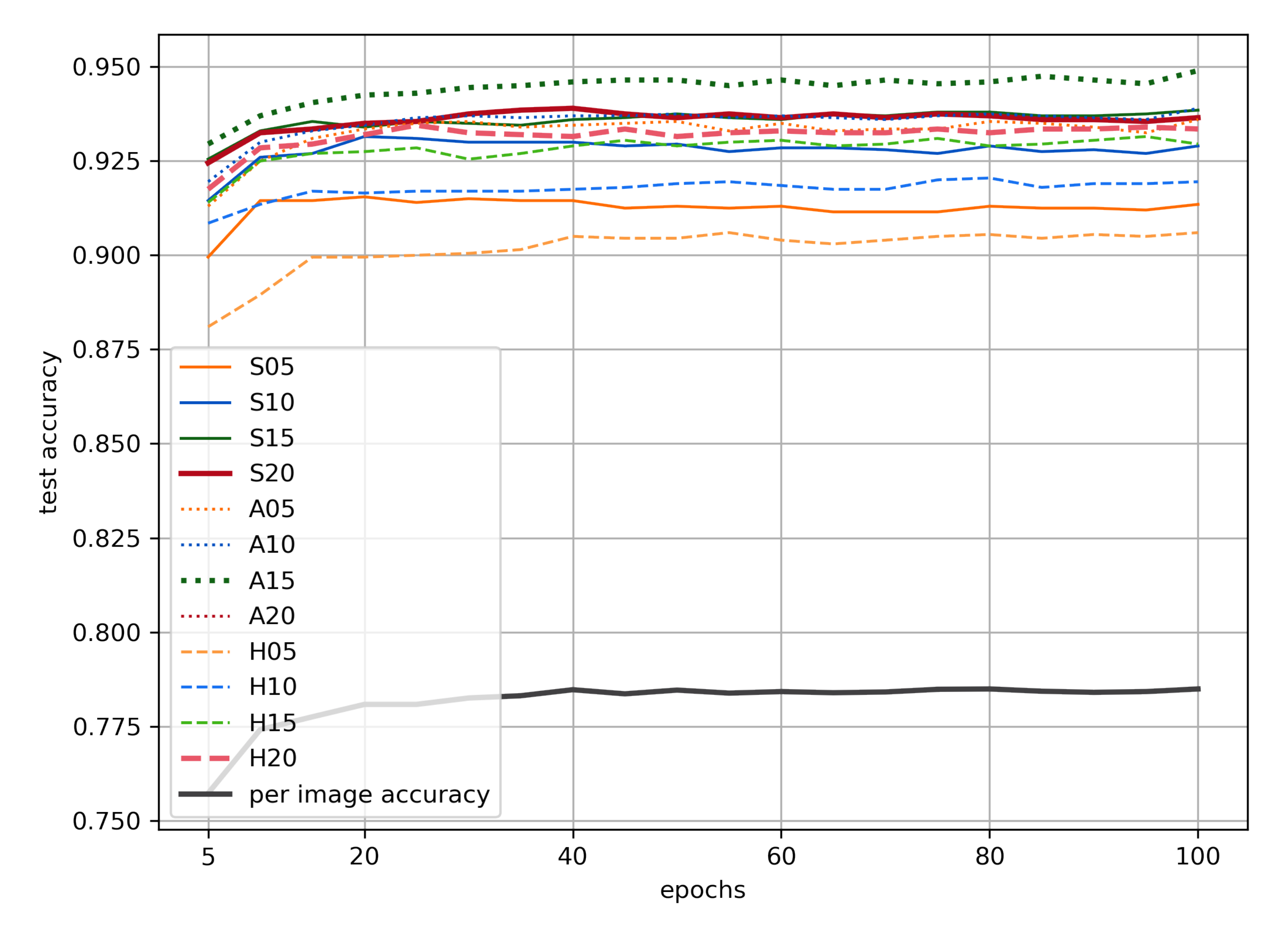}
    \caption{Success values of 13 accuracy metrics for the best experiment (Google10 \& DenseNet169) calculated on every 5 epochs.}
    \label{fig:metrics}
\end{figure*}

\begin{figure*}[t]
    \centering
    \includegraphics[width=\textwidth]{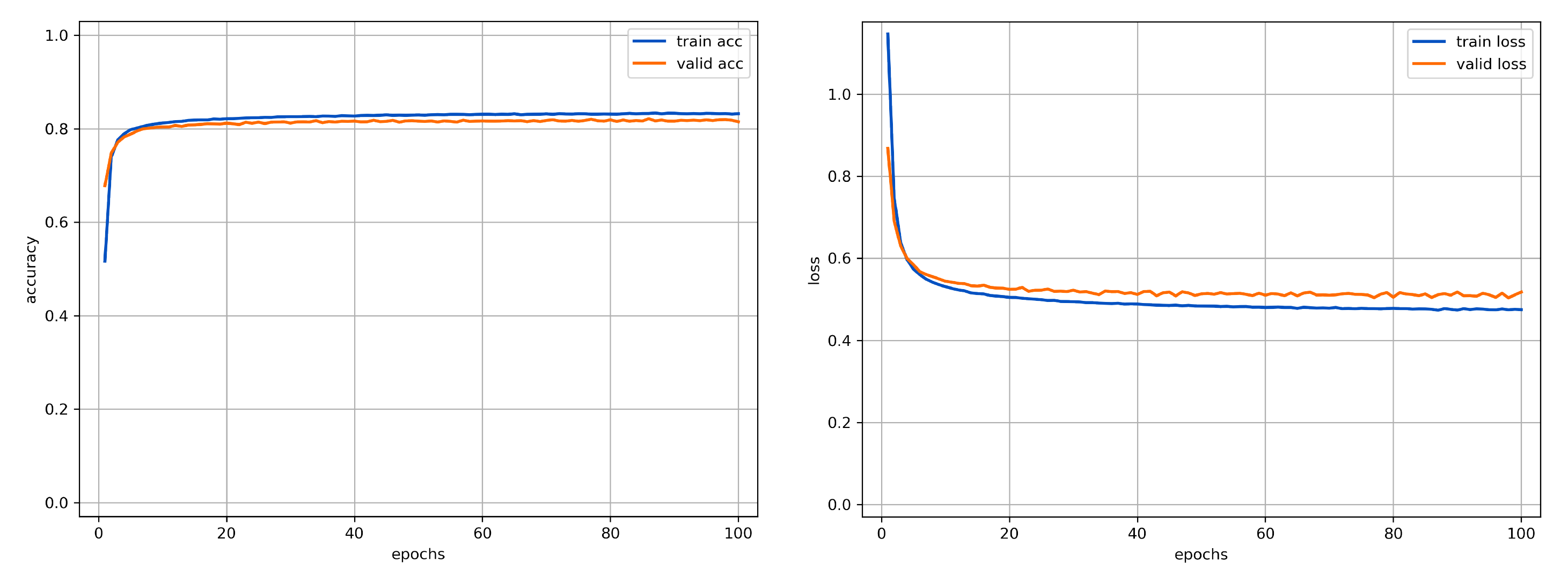}
    \caption{Accuracy and loss graphs of the best experiment (Google10 \& DenseNet169).}
    \label{fig:accloss}
\end{figure*}

We mentioned that our method had 12 metrics (S05, S10, S15, S20, H05, H10, H15, H20, A05, A10, A15, A20). We captured the maximum accuracy with A15 metric. After the result matrix was reordered for the greater averaged column, this metric used the results of the first 15 images and discarded the least related 5 images. Even the A05 metric, which used the results of the first 5 images, had good accuracies. We have already stated that Google Image Search results are related to the search query itself, not related to the class of the web site on the search query. So with reordering, we determined the most useful images over 20 images. On Figure \ref{fig:metrics} all metrics can be seen for best experiment: Google10 set with DenseNet169 model. All metrics run in every five epochs for the web pages in the test set. The important metrics are shown by ticker lines. Accuracy and loss graphs for the best experiment are shown in Figure \ref{fig:accloss}.

The used CNN models require square inputs with the size of 244x244 pixels. If there is an image in the dataset which has a different size, it is automatically resized to 244x244 pixels. There are 380318 images in Google Image Search results, and the histogram of the width-height ratio of these images (Figure \ref{fig:whr}) shows that most of these images are far from being square. Besides this, since they are thumbnails, they are small, only 30932 of these images have width and height larger than 224 pixels. These facts show that Google Image Search results are not the best input for our CNN models. But as mentioned, Google is a reliable source to obtain descriptive images for web pages and to find better inputs could be very difficult from other sources.

\subsection{Comparisons with other methods}

To compare our method with the other methods in the literature, we applied the other methods to the WebScreenshots dataset and discussed their successes on the same dataset that we used. WebScreenshots dataset contains one screenshot for each web site and also text contents of the web sites. We used these features for other methods and Google Image Search results for our method.

Videira and Goncalves used Bag of Visual Words (BoVW) to classify the topic of web pages \cite{videira2014automatic}. They achieved 82,50\% accuracy for 4 classes, with 90 images for each class, 360 images in total. When we applied this method to 20000 images (screenshots of WebScreenshots dataset), we only get 46.95\% accuracy. Our second experiment was to use BoVW in our method instead of CNN models. We trained subsets of WebScreenshots dataset (Subset10 and Subset20) using BoVW and Google Image Search results. After applying our metrics, we could only get 62.50\% accuracy at most.

BoVW accuracy was calculated with these steps: (1) RootSIFT descriptors \cite{arandjelovic2012three} were extracted from the images. (2) K-means clustering was performed to these descriptions. (3) TF-IDF values of the clusters were calculated. (4) The model was trained with linear SVM. After training, to find the class of a test web site, we performed the same processes with its screenshot or descriptive images and compared SVM results with trained SVM clusters.

RootSIFT descriptors have 128 dimensions, and clustering on multidimensional space has problems such as the curse of dimensionality. BoVW requires too much memory and time since it processes all input images at the same time. As a result, we concluded that BoVW is not a suitable method for large datasets. We could not run BoVW with Google10 and Google20 sets since it requires extreme memory resources for 380000 images. We tried using both SIFT and RootSIFT descriptors, and RootSIFT had slightly better results than SIFT.

Dou et al. classified web pages for their aesthetics value (binary classification) with deep learning methods \cite{dou2019webthetics}. Their dataset contained 398 web page screenshots, and they calculated the correlation of their deep learning with transfer learning model and actual user ratings as more than 90\%. For WebScreenshots dataset, all classes were predefined, and there were no user predictions. With this condition, when we applied similar methods to WebScreenshots dataset (20000 web sites and screenshots), we could get 68.40\% accuracy at most. On the other hand, our method achieved 97.50\% accuracy for 2000 and 94.90\% accuracy for 20000 web pages and 4 classes.

For the web classification problem, there are numerous researches which use text classification, since web sites have more textual features than visual features. To compare our method with textual classification methods, we contented with a quite common text classification method: Term Frequency - Inverse Document Frequency (TF-IDF). TF-IDF applied to web classification problem by Salamat and Omatu \cite{selamat2003neural}. They used a neural network with 25 hidden layers. Their dataset contains 1166 sports documents on 11 sports categories (baseball, boxing, tennis, etc.). They stated their success rate as 83.94\% for the TF-IDF method.

We applied TF-IDF on the WebScreenshots dataset. We used the text contents of web sites that came with the dataset. In our experiments, we tokenized the text contents and constructed TF-IDF matrices for training and test sets. We fed these matrices into a sequential deep learning network with 5 hidden layers. After trying different values for the vocabulary size parameter, we experienced a maximum 92.65\% success rate for TF-IDF on the WebScreenshots dataset. Adding more layers did not impact the result. Our descriptive images method had a higher success rate, 94.90\%, than text classification with TF-IDF and gave a better result for a multilingual dataset.

\section{Conclusion}\label{sec:conclusion}

In this paper, we have presented a web site classification system based on combining multiple neural network results into one. In the model, each web site represented by its descriptive images, which are the Google Image Search results for the web site URL address. A classification rate of 94.90\% was obtained when the model applied on a dataset of 20000 web sites. The main advantages of the proposed model are its mobility and its ease of application. The proposed model can be used for other problems where an element can be described with more than one feature. We used well known CNN architectures, 3 kinds of metrics, and Google Image Search results for our problem. But different or more CNN architectures, metrics, or descriptive images could be used according to the problem.

We also represent a new dataset, WebScreenshots, that contains 20000 web sites in 4 classes. The dataset contains screenshots and text contents of the web sites, and it is a multilingual dataset with 44 languages.

Since our model is a novel method, there are still some improvements to be considered for future works. We use predefined CNN architectures with transfer learning, so we had to use their default input size 224x224 pixels. Current input images could be filtered w.r.t. their weight-height ratio. In addition, the dataset can be expanded and experiments can be repeated for more classes. In this paper, we worked on a multi-class classification problem. There is a high probability that our method will perform better in binary classification problems, such as detecting adult web sites. We are also planning to improve the used CNN architectures and try new ones to get better results. At this point, we would like to mention the importance of the contributions of other researchers. As our method can be applied to different problems, we believe it will improve with each study.

\section*{Acknowledgements}

We like to thank to the authors of open source projects used in this work: Pavan Solapure for his text classifier example\footnote{https://www.opencodez.com/python/text-classification-using-keras.htm} (code revised to get maximum accuracy and to work with UTF-8 encoded files), Bikramjot Singh Hanzra for his bag of visual words example\footnote{https://github.com/bikz05/bag-of-words} (code upgraded to Python3 and RootSIFT added)

\begin{table}[th]
    \centering
    \caption{Sample result matrix for descriptive images of a web site}
    \label{tab:sampleresults}
    \small
    \begin{tabular}{l c c c c}
        \noalign{\smallskip}
        \hline
        \noalign{\smallskip}
        Image  & Machinery  & Music      & Sport      & Tourism    \\
        \noalign{\smallskip}
        \hline
        \noalign{\smallskip}
        01.jpg & 0.97641605 & 0.00005760 & 0.00185017 & 0.02167617 \\
        02.jpg & 0.99977142 & 0.00001599 & 0.00017324 & 0.00003941 \\
        03.jpg & 0.12737411 & 0.15132521 & 0.01362997 & 0.70767075 \\
        04.jpg & 0.39753565 & 0.16894254 & 0.06661761 & 0.36690423 \\
        ...    & ...        & ...        & ...        & ...        \\
        20.jpg & 0.99998045 & 0.00000056 & 0.00001787 & 0.00000108 \\
        \noalign{\smallskip}
        \hline
    \end{tabular}
\end{table}

\begin{table}[th]
    \centering
    \caption{One-hot transformation of the sample results on Table \ref{tab:sampleresults}}
    \label{tab:onehot}
    \small
    \begin{tabular}{lcccc}
        \noalign{\smallskip}
        \hline
        \noalign{\smallskip}
        Image  & Machinery  & Music      & Sport      & Tourism    \\
        \noalign{\smallskip}
        \hline
        \noalign{\smallskip}
        01.jpg & 1 & 0 & 0 & 0 \\
        02.jpg & 1 & 0 & 0 & 0 \\
        03.jpg & 0 & 0 & 0 & 1 \\
        04.jpg & 1 & 0 & 0 & 0 \\
        ...    & ... & ... & ... & ... \\
        20.jpg & 1 & 0 & 0 & 0 \\
        \noalign{\smallskip}
        \hline
    \end{tabular}
\end{table}

\begin{table}[th]
    \centering
    \caption{Average reordered transformation of the sample results on Table \ref{tab:sampleresults}}
    \label{tab:averagereordered}
    \small
    \begin{tabular}{lcccc}
        \noalign{\smallskip}
        \hline
        \noalign{\smallskip}
        Image  & Machinery  & Music      & Sport      & Tourism    \\
        \noalign{\smallskip}
        \hline
        \noalign{\smallskip}
        17.jpg & 0.99999964 & 0.00000003 & 0.00000033 & 0.00000001 \\
        16.jpg & 0.99999952 & 0.00000000 & 0.00000042 & 0.00000002 \\
        13.jpg & 0.99999654 & 0.00000323 & 0.00000011 & 0.00000008 \\
        14.jpg & 0.99999309 & 0.00000058 & 0.00000620 & 0.00000010 \\
        ...    & ...        & ...        & ...        & ...        \\
        03.jpg & 0.12737411 & 0.15132521 & 0.01362997 & 0.70767075 \\
        \noalign{\smallskip}
        \hline
    \end{tabular}
\end{table}

%
\section*{Conflict of interest}

The authors declare that they have no conflict of interest.



\begin{thebibliography}{10}
    
    \bibitem{ahmadi2011intelligent}
    Ali Ahmadi, Mehran Fotouhi, and Mahmoud Khaleghi.
    \newblock Intelligent classification of web pages using contextual and visual
    features.
    \newblock {\em Applied Soft Computing}, 11(2):1638--1647, 2011.
    
    \bibitem{arandjelovic2012three}
    Relja Arandjelovi{\'c} and Andrew Zisserman.
    \newblock Three things everyone should know to improve object retrieval.
    \newblock In {\em 2012 IEEE Conference on Computer Vision and Pattern
        Recognition}, pages 2911--2918. IEEE, 2012.
    
    \bibitem{arentz2004classifying}
    Will~Archer Arentz and Bj{\o}rn Olstad.
    \newblock Classifying offensive sites based on image content.
    \newblock {\em Computer Vision and Image Understanding}, 94(1-3):295--310,
    2004.
    
    \bibitem{aydos2020web}
    Fahri Aydos.
    \newblock Web{S}creenshots, https://www.kaggle.com/ds/202248, 2020.
    
    \bibitem{deboer2010classifying}
    Viktor de~Boer, Maarten van Someren, and Tiberiu Lupascu.
    \newblock Classifying web pages with visual features.
    \newblock In {\em Web Information Systems and Technologies}, pages 245--252,
    2010.
    
    \bibitem{deng2009imagenet}
    Jia Deng, Wei Dong, Richard Socher, Li-Jia Li, Kai Li, and Li~Fei-Fei.
    \newblock Imagenet: A large-scale hierarchical image database.
    \newblock In {\em 2009 IEEE conference on computer vision and pattern
        recognition}, pages 248--255. Ieee, 2009.
    
    \bibitem{dong2014adult}
    Kaikun Dong, Li~Guo, and Quansheng Fu.
    \newblock An adult image detection algorithm based on bag-of-visual-words and
    text information.
    \newblock In {\em 2014 10th International Conference on Natural Computation
        (ICNC)}, pages 556--560. IEEE, 2014.
    
    \bibitem{dou2019webthetics}
    Qi~Dou, Xianjun~Sam Zheng, Tongfang Sun, and Pheng-Ann Heng.
    \newblock Webthetics: quantifying webpage aesthetics with deep learning.
    \newblock {\em International Journal of Human-Computer Studies}, 124:56--66,
    2019.
    
    \bibitem{hammami2005webguard}
    Mohamed Hammami, Youssef Chahir, and Liming Chen.
    \newblock Webguard: A web filtering engine combining textual, structural, and
    visual content-based analysis.
    \newblock {\em IEEE Transactions on Knowledge and Data Engineering},
    18(2):272--284, 2005.
    
    \bibitem{hashemi2020web}
    Mahdi Hashemi.
    \newblock Web page classification: a survey of perspectives, gaps, and future
    directions.
    \newblock {\em Multimedia Tools and Applications}, pages 1--25, 2020.
    
    \bibitem{hashemi2019detecting}
    Mahdi Hashemi and Margeret Hall.
    \newblock Detecting and classifying online dark visual propaganda.
    \newblock {\em Image and Vision Computing}, 89:95--105, 2019.
    
    \bibitem{hu2007recognition}
    Weiming Hu, Ou~Wu, Zhouyao Chen, Zhouyu Fu, and Steve Maybank.
    \newblock Recognition of pornographic web pages by classifying texts and
    images.
    \newblock {\em IEEE transactions on pattern analysis and machine intelligence},
    29(6):1019--1034, 2007.
    
    \bibitem{huang2017densely}
    Gao Huang, Zhuang Liu, Laurens Van Der~Maaten, and Kilian~Q Weinberger.
    \newblock Densely connected convolutional networks.
    \newblock In {\em Proceedings of the IEEE conference on computer vision and
        pattern recognition}, pages 4700--4708, 2017.
    
    \bibitem{krizhevsky2012imagenet}
    Alex Krizhevsky, Ilya Sutskever, and Geoffrey~E Hinton.
    \newblock Imagenet classification with deep convolutional neural networks.
    \newblock In {\em Advances in neural information processing systems}, pages
    1097--1105, 2012.
    
    \bibitem{lecun1989backpropagation}
    Yann LeCun, Bernhard Boser, John~S Denker, Donnie Henderson, Richard~E Howard,
    Wayne Hubbard, and Lawrence~D Jackel.
    \newblock Backpropagation applied to handwritten zip code recognition.
    \newblock {\em Neural computation}, 1(4):541--551, 1989.
    
    \bibitem{mirdehghani2009web}
    Maryam Mirdehghani and S~Amirhassan Monadjemi.
    \newblock Web pages aesthetic evaluation using low-level visual features.
    \newblock {\em World Academy of Science, Engineering and Technology},
    49:811--814, 2009.
    
    \bibitem{rowley2006large}
    Henry~A. Rowley, Yushi Jing, and Shumeet Baluja.
    \newblock Large scale image-based adult-content filtering.
    \newblock In {\em International Conference on Computer Vision Theory and
        Applications}, volume~1, pages 290--296, 2006.
    
    \bibitem{russakovsky2015imagenet}
    Olga Russakovsky, Jia Deng, Hao Su, Jonathan Krause, Sanjeev Satheesh, Sean Ma,
    Zhiheng Huang, Andrej Karpathy, Aditya Khosla, Michael Bernstein, et~al.
    \newblock Imagenet large scale visual recognition challenge.
    \newblock {\em International journal of computer vision}, 115(3):211--252,
    2015.
    
    \bibitem{selamat2003neural}
    Ali Selamat and Sigeru Omatu.
    \newblock Neural networks for web page classification based on augmented {PCA}.
    \newblock In {\em Proceedings of the International Joint Conference on Neural
        Networks}, volume~3, pages 1792--1797, 2003.
    
    \bibitem{simonyan2014very}
    Karen Simonyan and Andrew Zisserman.
    \newblock Very deep convolutional networks for large-scale image recognition.
    \newblock {\em arXiv preprint arXiv:1409.1556}, 2014.
    
    \bibitem{dataDMOZ}
    Gaurav Sood.
    \newblock {Parsed DMOZ data}, 2016.
    
    \bibitem{szegedy2017inception}
    Christian Szegedy, Sergey Ioffe, Vincent Vanhoucke, and Alexander~A Alemi.
    \newblock Inception-v4, inception-resnet and the impact of residual connections
    on learning.
    \newblock In {\em Thirty-first AAAI conference on artificial intelligence},
    2017.
    
    \bibitem{videira2014automatic}
    Ant{\'o}nio Videira and Nuno Goncalves.
    \newblock Automatic web page classification using visual content.
    \newblock In {\em Web Information Systems and Technologies}, pages 193--204,
    2014.
    
\end{thebibliography}

\end{document}